\pdfoutput=1

\documentclass[11pt]{article}

\usepackage[]{ACL2023}

\usepackage{times}
\usepackage{latexsym}

\usepackage[T1]{fontenc}

\usepackage[utf8]{inputenc}
\usepackage{microtype}

\usepackage{inconsolata}

\usepackage{amsfonts} 
\usepackage{amssymb} 
\usepackage{pifont}
\usepackage{tabularx} 
\usepackage{arydshln} 
\usepackage{xcolor} 
\usepackage{mathtools} 
\usepackage{adjustbox} 
\usepackage{multirow} 
\usepackage{paralist} 
\usepackage{CJKutf8} 
\usepackage{booktabs} 
\usepackage{natbib} 
\usepackage{setspace}
\usepackage{enumerate}
\usepackage{bbding}
\usepackage{wasysym}

\usepackage{caption}
\usepackage{subcaption}
\usepackage{bbold}
\usepackage{tikz}
\usepackage{graphicx}
\usepackage{enumitem}

\usepackage[ruled,linesnumbered]{algorithm2e}
\SetKwInOut{Parameter}{Parameters}

\usepackage{amsthm} 
\theoremstyle{definition} 

\usepackage{tikz}
\usepackage[edges]{forest} 
\usepackage{pgfplots}
\usepackage{custom} 
\usepackage{amsmath}
\usepackage[british,american]{babel}
\usepackage{bm}
\usepackage{scalefnt}
\usepackage[most]{tcolorbox}

\usepackage{adjustbox}

\usepackage[edges]{forest}
\usepackage[framemethod=tikz]{mdframed}
\usepackage{subcaption}
\usepackage{soul}
\definecolor{lgreen}{rgb}{0.89,0.94,0.85}
\definecolor{lred}{rgb}{0.98, 0.90, 0.84}
\definecolor{lyellow}{rgb}{1.00, 0.95, 0.80}
\definecolor{lblue}{rgb}{0.85, 0.89, 0.95}
\definecolor{hidden-draw}{RGB}{20,68,106}
\definecolor{hidden-pink}{RGB}{255,245,247}

\newtcolorbox{disbox}[1][]{
  colback=white,
  colbacktitle=white,
  coltitle=black,
  fonttitle=\bfseries,
  bottomrule=0pt,
  toprule=0pt,
  leftrule=2pt,
  rightrule=2pt,
  titlerule=0pt,
  arc=0pt,
  outer arc=0pt,
  colframe=orange!60,
}

\tcbset{myprompt/.style={subtitle style={colback={blue3!20}}}}

\newtcolorbox{taskbox}[2][]{%
  enhanced, breakable,
  colframe=blue3!40,
  colback=blue5!5,
  arc=1mm,
  outer arc=1mm,
  fontupper=\small,
  fontlower=\small,
  coltitle=blue1,
  fonttitle=\bfseries,
  boxsep=1mm,
  left=0mm,
  right=0mm,
  top=0mm,
  bottom=0mm,
  before={\noindent},
  segmentation style={solid, blue3},
  title=#2,%
  #1
}

\definecolor{blue1}{rgb}{0.15,0.15,0.15}
\definecolor{blue2}{rgb}{0.30,0.30,0.30}
\definecolor{blue3}{rgb}{0.45,0.45,0.45}
\definecolor{blue4}{rgb}{0.60,0.60,0.60}
\definecolor{blue5}{rgb}{0.70,0.70,0.70}
\definecolor{blue6}{rgb}{0.80,0.80,0.80}

\newcommand*{\img}[1]{%
    \raisebox{-.3\baselineskip}{%
        \includegraphics[
        height=\baselineskip,
        width=\baselineskip,
        keepaspectratio,
        ]{#1}%
    }%
}

\newlength{\transcriptlen}

\NewDocumentCommand {\setspeaker} { mo } {%
  \IfNoValueTF{#2}
  {\expandafter\newcommand\csname#1\endcsname{\item[#1:]}}%
  {\expandafter\newcommand\csname#1\endcsname{\item[#2:]}}%
  \IfNoValueTF{#2}
  {\settowidth{\transcriptlen}{#1}}%
  {\settowidth{\transcriptlen}{#2}}%
}

\newcounter{chatno}

\newcolumntype{M}[1]{>{\centering\arraybackslash}m{#1}}
\newcolumntype{N}[1]{>{\centering\arraybackslash}c}

\title{A Preliminary Evaluation of ChatGPT for Zero-shot \\ Dialogue Understanding
}

\author{Wenbo Pan, Qiguang Chen$^{\ddag}$, Xiao Xu$^{\ddag}$, Wanxiang Che$^{\ddag}$, Libo Qin$^{\dag}$ \\
		$^\dag$School of Computer Science and
	Engineering,
	Central South University \\
	$^\ddag$Research Center for Social Computing and Information Retrieval \\
	$^\ddag$Harbin Institute of Technology, China \\
	
  pixelwenbo@gmail.com,
	\{qgchen, xxu, car\}@ir.hit.edu.cn, lbqin@csu.edu.cn
}

\begin{document}
\maketitle



\begin{abstract}

Zero-shot dialogue understanding aims to enable dialogue to track the user's needs without any training data, which has gained increasing attention. In this work, we investigate the understanding ability of ChatGPT for zero-shot dialogue understanding tasks including spoken language understanding (SLU) and dialogue state tracking (DST). Experimental results on four popular benchmarks reveal the great potential of ChatGPT for zero-shot dialogue understanding.
In addition, extensive analysis shows that ChatGPT benefits from the multi-turn interactive prompt in the DST task but struggles to perform slot filling for SLU.
Finally, we summarize several unexpected behaviors of ChatGPT in dialogue understanding tasks, hoping to provide some insights for future research on building zero-shot dialogue understanding systems with Large Language Models (LLMs).

\end{abstract}
\section{Introduction}
Recent studies on Large Language Models (LLMs), such as GPT-3~\cite{brown2020language}, InstructGPT~\cite{ouyang2022training}, PaLM~\cite{chowdhery2022palm}, and OPT~\cite{zhang2022opt}, have exhibited impressive zero-shot performance.
More recently, ChatGPT\footnote{We use the Jan 30 version of ChatGPT in this paper ( \href{https://chat.openai.com/chat}{https://chat.openai.com/chat}).}, a conversational large language model that has been trained by reinforcement learning with human feedback, which has brought
remarkable success on various zero-shot natural language processing (NLP) tasks.

Specifically, ChatGPT has shown competitive performance on
zero-shot logical reasoning~\cite{qin2023chatgpt}, text summarization~\cite{yang2023exploring}, machine translation~\cite{jiao2023chatgpt}, information extraction~\cite{wei2023zero} and so on.
However, it remains unclear how ChatGPT performs when it comes to dialogue-understanding tasks.

To this end, we provide an empirical analysis on performing ChatGPT for zero-shot dialogue understanding tasks including spoken language understanding (SLU) and dialogue state tracking (DST).
Furthermore, we introduce a multi-turn interactive prompt framework to improve the performance of ChatGPT in multi-turn DST.
We conduct experiments on four widely used benchmarks including ATIS~\cite{hemphill-etal-1990-atis}, SNIPS~\cite{couke2018snips} in SLU and MultiWOZ2.1~\cite{eric2019multiwoz}, MultiWOZ2.4~\cite{Ye2021MultiWOZ2A} in DST.

Through a preliminary exploration study, we provide the following observations:

\begin{itemize}
    \item[\img{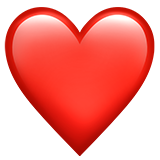}]ChatGPT demonstrates better capability in multi-turn dialogue understanding tasks (multi-turn DST) as compared to single-turn tasks (single-turn SLU).
    \item[\img{figures/heart-emoji.png}] Multi-turn interactive prompts can better leverage ChatGPT's multi-turn ability to enhance multi-turn tasks (i.e., DST).
    \item[\img{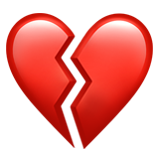}] ChatGPT exhibits relatively inferior performance in slot-filling tasks, which can be compensated for by providing descriptions and examples of slot names.
    \item[\img{figures/broken-heart-emoji.png}] During multi-turn conversations, ChatGPT may occasionally exhibit unexpected behaviors, such as violating format requirements.
\end{itemize}

\label{Introduction}

\section{ChatGPT for Dialogue Understanding}

\begin{table}[h]
	\centering
	\begin{adjustbox}{width=0.43\textwidth}
		\begin{taskbox}[myprompt]{Example of SLU task}
			\vspace*{-0.12cm}
			
			\tcbsubtitle{Schema and Regulations}
			
			\texttt{\textcolor{blue3}{[Intent constraints]}}
			
			Given the following sentences, first choose the intent of the sentences from the following intent list: [...] \\

			\texttt{\textcolor{blue3}{[Slot constraints]}}

			And then annotate given sentences with slots from the following slot list, and the sample values of each slot are given. [album: Like A Hurricane, The Happy Blues ...] \\

			\texttt{\textcolor{blue3}{[Regulations]}}

			You need to output the annotations in the form of "..." \\			
			You must not output anything other than the annotations... \\
			You must not miss any possible slot-value pairs...
			\tcbsubtitle{Sentence input}
			\texttt{put United Abominations onto my rare groove playlist:}
			\tcbsubtitle{ChatGPT response}
			The intent is AddToPlaylist; United Abominations is an "entity-name" entity; ...
			\tcbsubtitle{Parsed answer}
			intent=\texttt{AddToPlaylist}, \texttt{entity-name}="United Abominations"
		\end{taskbox}
	\end{adjustbox}
	\caption{Example of prompt for SLU.}
	\label{tab:snips_example}
\end{table}
\subsection{ChatGPT for Zero-Shot SLU}
SLU typically contains two sub-tasks including intent detection and slot filling, which are used to extract the user's slots and intents~\cite{ijcai2021p622}.

In this section, we directly explore ChatGPT with prompts for zero-shot SLU. As shown in Table~\ref{tab:snips_example},
the prompt contains three parts: schema, regulations and sentence input:
\begin{itemize}
	\item \textbf{Schema} aims to provide additional information on SLU for ChatGPT. It consists of intent constraints and slot constraints to encourage ChatGPT to generate correct intents and slots. Specifically, the intent constraint is a list of all the possible intents that ChatGPT can use,  while slot constraints provide examples of valid values and descriptions for each slot.
	\item \textbf{Regulations} are used to guide ChatGPT to generate reasonable responses.
	      As shown in Table~\ref{tab:snips_example}, we require ChatGPT to first predict intent with template \texttt{"The intent is <intent>"}. Then, all extracted slot-value pairs are restricted in form of \texttt{"The <value> is an <slot> entity;"}.
	      We also instruct ChatGPT not to generate anything other than the above valid outputs.
	\item \textbf{Sentence Input} is the samples used for testing. Give the input in Table~\ref{tab:snips_example} as an example, we ask ChatGTP to predict the corresponding intents and slots of sentence input \textit{"put United Abominations onto my rare groove playlist:"},
\end{itemize}

\subsection{ChatGPT for Zero-Shot DST}
\begin{table}[t]
	\centering
	\begin{adjustbox}{width=0.43\textwidth}
		\begin{taskbox}[myprompt]{Example of multi-turn DST}
			\vspace*{-0.12cm}
			\tcbsubtitle{Schema and Regulations}
			\texttt{\textcolor{blue3}{[Table descriptions]}}

			I present you with some databases together with example items and value constraints\\
			\texttt{Hotel} \\
			\texttt{|Name\ \ \ \ |type\ \ |parking |area\ \ |}  \\
			\texttt{|lodge\ \ \ |guest |yes\ \ \ \ \ |north\ |}  \\
			\texttt{|hamilton|house |no\ \ \ \ \ \ |east\ \ |}  \\
			where parking can be one of [free, yes, no] ... \\

			\texttt{\textcolor{blue3}{[Regulations]}}

			Your need to output in the following form for every turn I give you:

			User informed 'N' columns: <table name> - <column> = <value>; ...
			
			You must output attributes that are valid in the above tables

			You shouldn't output any attributes that have already appeared in any previous lists.

			If and only if the user explicitly said he/she doesn't care about some attribute, you should output '<dontcare>' for that attribute.
			... 

			\tcbsubtitle{Sentence input \#1}
			Next is the first turn:\\
			\textit{System:}\\
			\textit{User: I would like a taxi from Saint johns College to pizza hut fen ditton}
			\tcbsubtitle{ChatGPT response \#1}
			User informed 2 columns: tsynaxi-dep	[arture=saint john s college; taxi-destination=pizza hut fen ditton;

			\tcbsubtitle{Sentence input \#2}
			2th turn: \\
			\textit{System: what time do you want to leave and what time do you want to arrive by?}\\
			\textit{User: I want to leave after 17:15.}
			\tcbsubtitle{ChatGPT response \#2}
			User informed 1 column: taxi-leaveat=after 17:15;
		\end{taskbox}
	\end{adjustbox}
	\caption{Examples of multi-turn interactive prompt for DST.}
	\label{tab:mwz_example}
\end{table}

DST aims to track the user's needs at each turn when chatting with users.
In this section, we explore how to use ChatGPT for zero-shot DST~\cite{Jacqmin2022DoYF}.

\begin{table*}[h]
	\centering
	\begin{adjustbox}{width=\textwidth}
		\begin{tabular}{lcccccccc}
			\toprule
			\multirow{2}{*}{\textbf{Model}} & \multicolumn{2}{c}{SNIPS} & \multicolumn{2}{c}{ATIS} & \multicolumn{2}{c}{MultiWOZ2.1} & \multicolumn{2}{c}{MultiWOZ2.4}                                                       \\
			\cmidrule(r){2-3}
			\cmidrule(r){4-5}
			\cmidrule(r){6-7}
			\cmidrule(r){8-9}
			                                & Intent                    & Slot                     & Intent                          & Slot                            & JGA   & Slot Accuracy   & JGA   & Slot Accuracy   \\
			\midrule
			GPT-3.5                         & 98.00                     & 68.90                    & 90.03                           & 55.72                           & 32.25 & 94.79  & 34.55 & 95.21  \\
			Codex                           & 98.42                     & 68.90                    & 89.92                           & 57.29                           & 34.38 & 95.12     & 37.50 & 95.68    \\
			Finetuned SoTA                  & \textbf{98.80}                     & \textbf{97.10}                    & \textbf{98.00}                           & \textbf{96.10}                           & \textbf{61.02} & \textbf{98.05}      & \textbf{75.90} & -           \\
			\midrule
			ChatGPT                         & 97.71                     & 58.24                    & 75.22                           & 15.71                           & 60.28 & 97.83   & 64.23 & \textbf{98.12}   \\
			\bottomrule
		\end{tabular}
	\end{adjustbox}
	\caption{Results of zero-shot SLU and DST benchmarks. For the SLU benchmarks, Intent accuracy and Slot F1 are adopted as metrics. For the MultiWOZ datasets, we employ the Joint Goal Accuracy (JGA) and Slot Accuracy metrics as evaluation metrics. The results of ATIS, SNIPS, MultiWOZ2.1 and 2.4 are obtained from previous studies \citep{qin2021co,guo-etal-2022-beyond, zhao2022description}, respectively. ``-'' indicates the original paper does not
		report results.  }
	\label{tab:slu_results}
\end{table*}
Recent studies have indicated that the performance of ChatGPT may be suboptimal and prone to hallucination when presented with single-turn prompts~\cite{bang2023multitask}. To address this issue, we introduce a multi-turn interactive prompt approach for ChatGPT, which allows the model to track dialogue states during ongoing conversations.
An example of our DST prompt is shown in Table~\ref{tab:mwz_example} and the process description is illustrated as:
\begin{itemize}
	\item \textbf{In the first prompt turn}, we provide ChatGPT with a database description. In contrast to previous approaches that use SQL expressions~\cite{hu2022context}, we combine an example table in relational database format with natural language descriptions to create a more intuitive and easily understandable prompt.
	\item \textbf{In the subsequent turns}, we present each dialogue turn, including both system and user utterances, to ChatGPT one by one. ChatGPT is then required to generate the updated dialogue state for the current turn, based on the previously predicted dialogue states.
	      By doing this, ChatGPT can make full use of the previous dialogue contexts.
\end{itemize}

The multi-turn interactive prompt can improve ChatGPT's performance and reduce its tendency to generate irrelevant or inaccurate responses.
\section{Experiments}
\subsection{Datasets \& Metrics}
For zero-shot SLU, we use the test of ATIS~\cite{hemphill-etal-1990-atis}  and SNIPS~\cite{couke2018snips} to evaluate the zero-shot SLU performance.
For evaluating zero-shot DST performance, we use the test set of MultiWOZ2.1~\cite{eric2019multiwoz} and MultiWOZ2.4~\cite{Ye2021MultiWOZ2A} datasets.

\subsection{Baselines}
We compare ChatGPT with the following large language models:
(1) \textbf{GPT-3.5}~\cite{brown2020language,ouyang2022training} is a language model with 175B parameters that have been pre-trained on an extensive web corpus.
In this paper, we use \textbf{text-davinci-003} version of GPT-3.5 from OpenAI API.
(2) \textbf{Codex}~\cite{Chen2021EvaluatingLL} is another large language model trained on open-source code on GitHub.

In addition, we also report the results of recent state-of-the-art models on the four benchmarks to provide a comparative analysis.
Specifically, we choose the model proposed by \citet{qin2021co} for SLU. In DST, we adopt the results used by \citet{guo-etal-2022-beyond} and \citet{zhao2022description} for MultiWOZ2.1 and MultiWOZ2.4, respectively.

\subsection{Main Results}
The main results are illustrated in Table~\ref{tab:slu_results}.
We have the following observations:
\begin{itemize}
	\item [(1)] ChatGPT can achieve zero-shot dialogue understanding tasks. Although there is a gap between the results and fine-tuned SOTA, exploring ChatGPT is still a meaningful direction for zero-shot dialogue understanding task.
	\item [(2)] ChatGPT surpasses GPT-3.5 and Codex on MultiWOZ2.1 and MultiWOZ2.4. We attribute it to the fact that
	the proposed multi-turn interactive prompts can better leverage ChatGPT's multi-turn ability to improve DST performance.
	\item [(3)] The performance of ChatGPT underperforms ChatGPT and Codex on SLU benchmarks, which suggests that the performance of ChatGPT in slot filling task is sub-optimal and is consistent with the recent observation~\cite{qin2023chatgpt}.
\end{itemize}
\subsection{Analysis}

\begin{table}[t]
  \centering
  \begin{adjustbox}{width=0.5\textwidth}
    \begin{tabular}{lccc}
      \toprule
      \multirow{2}{*}{\textbf{Model}} & \multicolumn{3}{c}{MultiWOZ2.1}                      \\
      \cmidrule(r){2-3}
                                      & JGA                             & Slot Accuracy  \\
      \midrule
      Multi-turn interactive prompt   & 60.02                           & 97.80      \\
      Single-turn prompt      & 58.05                           & 97.74    \\
      \bottomrule
    \end{tabular}
  \end{adjustbox}
  \caption{Multi-turn interactive prompt vs. Single-turn prompt.}
  \label{tab:multi-single-turn}
\end{table}

\begin{figure*}[htbp]
  \centering
  \begin{minipage}[t]{0.32\textwidth}
    \begin{taskbox}[myprompt]{Undefined slot values}
      \vspace*{-0.12cm}
      \tcbsubtitle{Utterance}
      No preference, please just pick 1 and give me the postcode and address.
      \tcbsubtitle{Gold states}
      \texttt{attraction-area=dontcare}
      \tcbsubtitle{Prediction}
      \texttt{attraction-area=dontcare; attraction-name=\textcolor{red}{<unknown>}\\}
    \end{taskbox}
  \end{minipage}
  \hfill
  \begin{minipage}[t]{0.32\textwidth}
    \begin{taskbox}[myprompt]{Slot format violation}
      \vspace*{-0.12cm}
      \tcbsubtitle{Utterance}
      I want the train to leave at 5 p.m.
      \tcbsubtitle{Gold states}
      \texttt{train-leaveat=5 p.m.}
      \tcbsubtitle{Prediction}
      \texttt{train-leaveat=\textcolor{red}{"at 5 p.m."}\\\\}
    \end{taskbox}
  \end{minipage}
  \hfill
  \begin{minipage}[t]{0.32\textwidth}
    \begin{taskbox}[myprompt]{Verbose response}
      \vspace*{-0.12cm}
      \tcbsubtitle{Utterance}
      Book me some rooms to accommodate 8 people and provide me with the reference number.
      \tcbsubtitle{Gold states}
      \texttt{hotel-book-people=8}
      \tcbsubtitle{Prediction}
      \texttt{\textcolor{red}{I am not able to provide a booking reference number for} \textcolor{red}{rooms to accommodate 8 people as I am just an AI language model...}}
    \end{taskbox}
  \end{minipage}
  \caption{Typical unexpected behaviors of ChatGPT on dialogue understanding. }
  \label{fig:slu_error}
\end{figure*}

\subsubsection{Multi-Turn Interactive Prompts Can Perform Better DST}
To evaluate the effectiveness of the proposed multi-turn interactive prompts for DST, we directly use a single-turn prompt to predict the DST results where each unique ChatGPT dialogue session predicts the dialogue states of each turn.
The comparison results are shown in Table~\ref{tab:multi-single-turn}. We observe that the multi-turn prompt surpasses the single-turn prompt on all metrics.
We attribute it to the fact that multi-turn interactive prompts can better leverage ChatGPT's multi-turn ability to improve DST performance.

\subsubsection{Additional Information (Description and Example) can Boost Zero-Shot SLU Performance}

We evaluate the effectiveness of providing slot names only (Name Only), slot descriptions (w/ Des.), examples (w/ Exp.), or a combination of all three.

The experimental results are presented in Table~\ref{tab:slu_com}. The findings suggest that providing both slot names and descriptions leads to the best performance of slot filling on the overall metric, indicating the importance of providing relevant information. 

\begin{table}[t]
  \centering
  \begin{tabular}{lccc}
    \toprule
    \multirow{2}{*}{\textbf{Model}} & \multicolumn{3}{c}{SNIPS}         \\
    \cmidrule(r){2-4}
                                    & Intent                    & Slot & Overall  \\
    \midrule
    Name only.                      & 96.43                     & 25.78 & 9.29 \\
    w/ Description.                 & 96.00                     & 33.88 & 12.57 \\
    w/ Example.                     & 97.71                     & 58.24 & 28.86\\
    w/ Des+Exp.                     & 96.00                     & 59.08 & 26.29\\
    \bottomrule
  \end{tabular}
  \caption{Impact of Prompt Design on SLU Performance of ChatGPT. }
  \label{tab:slu_com}
\end{table}

\subsubsection{Unexpected Behaviors of ChatGPT}

ChatGPT demonstrates some undesired behaviors that may prevent the correct parsing of output. We summarize these behaviors into three categories, which are shown in Figure~\ref{fig:slu_error}:

\begin{enumerate}
  \item  \textbf{Undefined Slot Values}: ChatGPT output slot values with special tokens (e.g., \texttt{unknown, request}) to indicate constraints on slot names that are not required under our DST settings. As shown on the left of Figure~\ref{fig:slu_error}, ChatGPT generates the slot value \texttt{<unknown>} for the \texttt{attraction-name} slot incorrectly.
  
  \item \textbf{Slot Format Violations}: Some outputs violate our format requirements. Take the prediction in Figure~\ref{fig:slu_error} as an example, ChatGPT predicts \texttt{at 5 p.m.} as the value for slot \texttt{train-leaveat}, whereas the correct format for a time expression should not contain prepositions.
  \item \textbf{Verbose Responses}: For scenarios where it is difficult to provide a correct answer, ChatGPT may use natural language as the answer instead of formatted slot-value pairs. An example of a verbose output is illustrated on the right side of Figure~\ref{fig:slu_error}.
\end{enumerate}
\paragraph{Prompt Length Limitation}
While the multi-turn interactive prompt framework has successfully reduced prompt complexity and improved the performance of DST, we identify a forgetting problem that occurs during some long conversations (more than 10 turns) due to the length limitation of ChatGPT. Specifically, after several turns, ChatGPT may forget the first prompt.

\section{Conclusion}
\label{conclusion}
In this paper, we investigated ChatGPT for zero-shot dialogue understanding. We optimized prompt design and proposed an interactive multi-turn prompt framework to improve ChatGPT's performance. 
Experimental results demonstrated the great potential of ChatGPT for zero-shot dialogue understanding tasks. 

\section*{Limitations and Future work}
There are several limitations in this version, which can be improved in future work.
\begin{itemize}
	\item \textbf{More LLM Baselines} In the future, we can include more LLM baselines to give a thorough comparison of different large language models (LLMs) for zero-shot dialogue understanding tasks.
	\item \textbf{More Scenarios.} This version does not cover all zero-shot scenarios such as zero-shot cross-domain~\cite{liu-etal-2020-coach} or zero-shot cross-lingual SLU and DST~\cite{ijcai2020p533}.
	\item \textbf{Results update.} Since the ChatGPT model is constantly updating, we can update  the observation conclusions according to the new experimental results in the future.
\end{itemize}

\bibliographystyle{acl_natbib}

\bibliography{custom.bib}
\appendix
\clearpage

\end{document}